# Context aware saliency map generation using semantic segmentation

Mahdi Ahmadi
Department of Electrical and
Computer Engineering, Isfahan
University of Technology,
Isfahan, Iran 84156-83111
mahdi.ahmadi@ec.iut.ac.ir

Mohsen Hajabdollahi
Department of Electrical and
Computer Engineering, Isfahan
University of Technology,
Isfahan, Iran 84156-83111
m.hajabdollahi@ec.iut.ac.ir

Nader Karimi
Department of Electrical and
Computer Engineering, Isfahan
University of Technology,
Isfahan, Iran 84156-83111
nader.karimi@cc.iut.ac.ir

Shadrokh Samavi
Department of Electrical and
Computer Engineering, Isfahan
University of Technology,
Isfahan, Iran 84156-83111
samavi96@cc.iut.ac.ir

*Abstract*— **Saliency map detection, as a method for detecting important regions of an image, is used in many applications such as image classification and recognition. We propose that context detection could have an essential role in image saliency detection. This requires extraction of high level features. In this paper a saliency map is proposed, based on image context detection using semantic segmentation as a high level feature. Saliency map from semantic information is fused with color and contrast based saliency maps. The final saliency map is then generated. Simulation results for Pascal-voc11 image dataset show 99% accuracy in context detection. Also final saliency map produced by our proposed method shows acceptable results in detecting salient points.**

## I. INTRODUCTION

Image saliency is introduced as a metric for image importance analysis. Image saliency represents importance of the image regions for the human visual system. The eye fixation data is used to create the image saliency ground-truth [1]. Eye tracking methods are applied on different people to measure how much attention they have on each region of an image [1].

Image saliency analysis is widely used in many applications such as video compression [2], object recognition [3], photo rearrangement [4] and image classification [5]. Although Image saliency is studied in many recent researches but its accuracy still is a challenging problem. One of the most important factors to determine the pixel saliency is its contrast in its neighboring area. Considering the relationship between local image contrast and other regions give better insight in contrast formation. Hence, in [6] the structure contrast with an energy function minimization is used to create contrast measure suitable for saliency map. Image contrast, by itself cannot be a proper feature for saliency map generation and must be used along with other features to efficiently generate saliency map. Color specifications can be considered as another factor to make attention in human visual system.

There are many recent studies on saliency detection concentrating on the image color specifications. Global color information as well as local image information in neighboring area are considered as efficient features for saliency detection. In [7], a high dimensional color transform and random forests are used to extract saliency map. In some of recent studies both of the contrast and color features are used for saliency map generation. In [8] local and global information are used simultaneously for saliency map generation. Results of color and contrast features are then fused using Principle Component Analysis (PCA). In [9] the homogeneity of the saliency results are improved by utilizing a segmentation method as a means for saliency detection.

In recent years Convolutional Neural Networks (CNNs) are widely used is most computer vision applications. CNNs are introduced as the state of the art method in classification, pattern recognition and image segmentation. CNNs are also widely used as powerful semantic segmentation tools. Semantic data is considered as one of the high-level information which an image holds and can be used as output for the CNNs. On the other hand, learning methods on CNNs can be used directly in saliency map detection [10].

In this paper semantic data using a trained CNN as well as color and contrast data are utilized to generate the saliency map. At first, importance level of each pixel of the image is found based on the image context. Then saliency map is created using semantic segmentation. CNN produces both the semantic segmentation and the image context. Two other saliency maps are also generated based on color and contrast. Finally saliency map generated from the three mentioned methods are fused and the final saliency map is generated. Using high level features, e.g. semantic segmentation in saliency detection, improves the homogeneity as well as fidelity of the results.

## II. SEMANTIC SEGMENTATION AND CONTEXT DETECTION

Semantic segmentation can be used as high level feature for saliency map detection. Recently Fully Convolutional Networks (FCNs) have been used as effective method to most of computer vision tasks. We will show that the order of the image regions based on their semantic segmentation can be useful for saliency detection. In the followings the FCN structure for semantic segmentation and context detection are briefly reviewed.

## A. Fully Convolutional neural Network

In recent years, CNNs are extensively used in previously challenging tasks such as object detection, image classification and scene understanding. Three most important parts of the network structure include convolutional layer, pooling layer, and fully connected layer. Each convolutional layer contains convolution operators and a non-linear activation function. Convolutional layers are used for efficient feature extraction and as the number of convolutional layers is increased, the network can be used to approximate more complex non-linear functions. Due to the efficient network structure, the number of network parameters is reduced dramatically and its training process and implementation are simplified. Because of the local filtering in convolution layers, the problem of over-fitting is reduced. In recent years Fully Convolutional Networks (FCNs) are introduced as a version of conventional CNNs capable for performing semantic segmentation. A sample FCN structure is illustrated in Fig. 1 [11]. FCN structure is similar to CNN structures except the fully connected layers are replaced with up-sample layers.

## B. Image context

Similar objects in different environments can be meant in different ways, so human visual attention is different in different environments. For example attention to a building in a jungle view can be different to the same building but in a city view. In this regard *higher* or *lower* attention on different objects can be considered in different applications. Image view in which different objects result different meanings can be considered as image context. Image context detection plays an important role to determine the importance of the image objects. For example, the context of the traffic control system images can be considered as vehicle. In this context vehicles are the most important objects and buildings are not important at all. Image context detection can be considered as an effective preprocessing step for saliency detection.

## III. PROPOSED METHOD

In this section our proposed method for saliency detection is presented. The block diagram of the proposed method is illustrated in Fig. 2. It can be observed from Fig. 2 that saliency map is generated based on three features, which are color, contrast, and semantic segmentation. In the followings the proposed method is presented in more details.

## A. Saliency maps by color and contrast

Since human cognition system can process a limited data at a time, its focus would be on certain regions determined by the color and contrast of regions [5]. Regions with higher contrast can be better visualized. Therefore, converting the image from RGB to HSV can be useful for extraction of the image color and value.

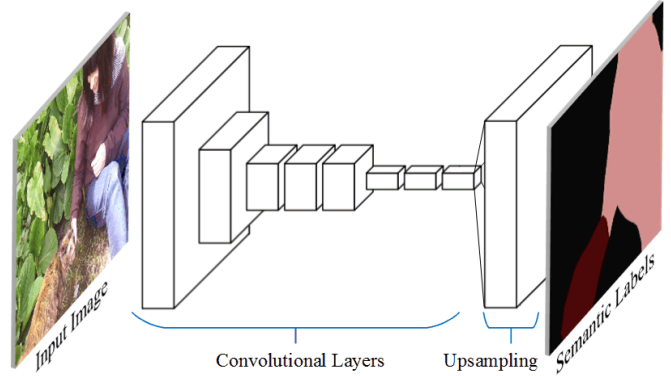

Figure. 1. Fully Convolutional Neural Network Structure [11]

Local pixel contrast is calculated based on the energy of the surroundings of an image block. Energy of each block in channel $V$ is obtained by (1) where $V_i$s are pixel values and $\overline{V}$ is the mean value of the pixels of the block and $S_{CN}$ is the contrast saliency.

$$S_{CN} = \frac{1}{n}\sum_{i=1}^{n}(V_i - \overline{V})^2 \qquad (1)$$

In the color selection, the higher saliency values are assigned to the warmer colors. For this purpose, a spectral filtering is applied on the hue channel in HSV space as described by (2), where $H$ is the magnitude of this channel and $S_{CL1}$ is the color saliency. Based on (2), a saliency value from zero (least significant color) to one (most significant color) is obtained. For adaptive filtering a parameter $P$ is defined as (3) in which higher $P$ makes the filter sharper.

$$S_{CL1} = \frac{1}{2}(cos(2\pi H) + 1) \qquad (2)$$

$$S_{CL} = S_{CL1}^{P} \qquad (3)$$

Fig. (3) illustrates applying spectral filter on Channel V to reach color saliency.

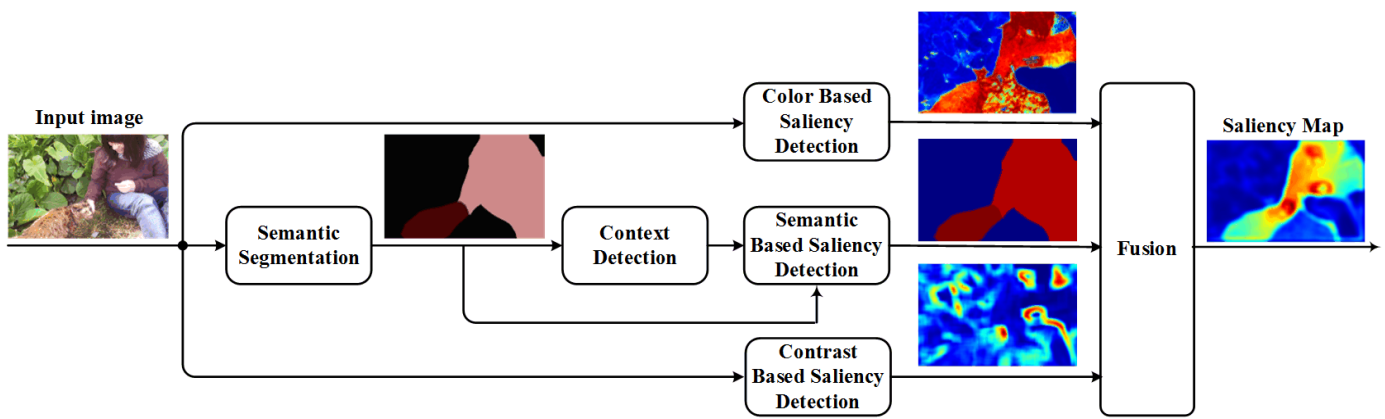

Figure. 2. Schematic diagram of the proposed method

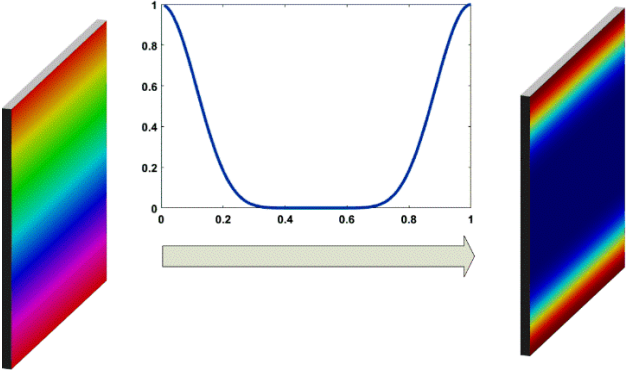

Figure. 3 Spectral filtering for color saliency.

### B. Context detection

Semantic segmentation results can be used as a high level feature for context detection. For this reason, we classified segmentation results into five classes as shown in Table 1. Then a multilayer neural network is applied for classification as depicted in Fig. 4. The area of each segmented object is used as input feature of the neural network. Finally, output layer of the classifier is equipped with a soft-max layer. The same training image set that is used for semantic segmentation is also used in case of the context detection.

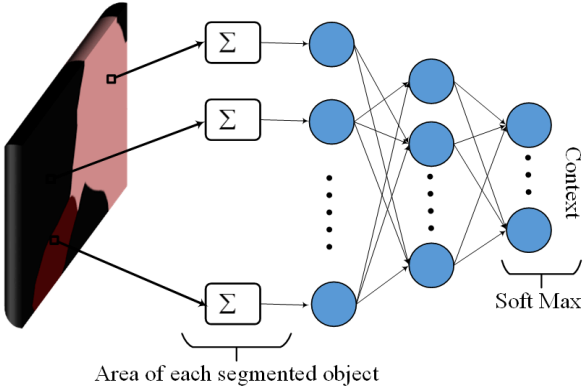

Figure. 4. context detection

Table 1. Image contexts of the proposed method.

| context | Pet | Other animals | Vehicle | Indoor | Others |
|---|---|---|---|---|---|
| Number of training images | 243 | 284 | 549 | 236 | 152 |
| Number of test images | 240 | 297 | 537 | 227 | 148 |

### C. Semantic based saliency map

The semantic segmentation label for each pixel has the most important information for the saliency detection map. To generate saliency map from pixel labels, a Look-Up Table (LUT) structure is used as shown in Fig. 5. Level of importance of each pixel is determined by an LUT. This importance level is different in different LUTs based on the contexts of an image. There are six LUTs which five of them are based on the image context and one LUT can be user defined depending on the needs of the user.

As illustrated in Fig. 5, each pixel semantic label is feed to all six LUTs and six corresponding saliency maps are generated. Finally, saliency map is selected from the LUT outputs based on pixel context and application.

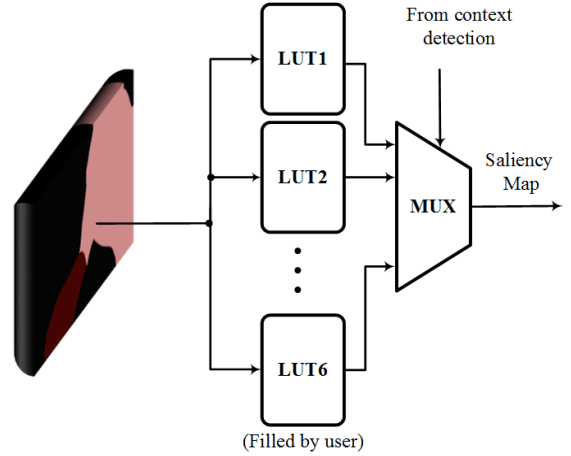

Figure. 5. Semantic based saliency map.

### D. Saliency map fusion

Three saliency maps, based on image semantics, color, and contrast, are combined with each other using a fusion method. At first, all three saliency maps are normalized and color based saliency map ($S_{CL}$) is combined with contrast based saliency map ($S_{CN}$) by a linear function using (4). $\omega_1$ and $\omega_2$ are linear coefficients which determine the importance of each map in the final saliency map and $S_{CN,CL}$ is color-contrast saliency map.

$$S_{CN,CL} = \omega_1 S_{CN} + \omega_2 S_{CL} \qquad (4)$$

Final saliency map ($S_{final}$) is obtained using (5) in which $S_{SSEG}$ is saliency based on semantic segmentation.

$$S_{final} = S_{SSEG} \cdot S_{CN,CL} \qquad (5)$$

### E. post processing

Central regions of image are more noticeable for human eyes. In this regard the saliency map resulted from the previous section is modified using a two dimensional Gaussian function as (6). $D$ is the Euclidean distance between the central image pixel and each pixel with coordinates $(x, y)$ that is obtained by (7). Also, $M$ and $N$ are image dimensions and $\sigma^2$ is the variance of the Gaussian function.

$$f(x,y) = 2 + e^{-\frac{D^2}{\sigma^2 \max(M,N)}} \qquad (6)$$

$$D = \sqrt{\left(x - \frac{M}{2}\right)^2 + \left(y - \frac{N}{2}\right)^2} \qquad (7)$$

Finally, the saliency map is smoothed with a low-pass filter for better visual quality.

## IV. EXPERIMENTAL RESULTS

Saliency detection algorithm is trained and tested on Pascal-voc11 image dataset [12]. Pascal-voc11 data set contains 2913 images which are semantically segmented into 20 classes. The class information for segmentation contest is included in Table 2. We used pre-trained network parameters for semantic segmentation [13]. All 20 object classes are considered to be a member of the five contexts of Table 1. Using Keras, an MLP structure is applied for context detection which has two hidden layers with 120 neurons. Relu is applied as the activation function and in the output layer a softmax is utilized. The parameter $P$ in (3) is considered to be equal to 4 and we experimentally set $\omega_1=\omega_2=0.5$ in (4). For post processing, we set $\sigma^2$ to 40 in (6) and for low-pass filtering a mean filter with size 20×20 is applied.

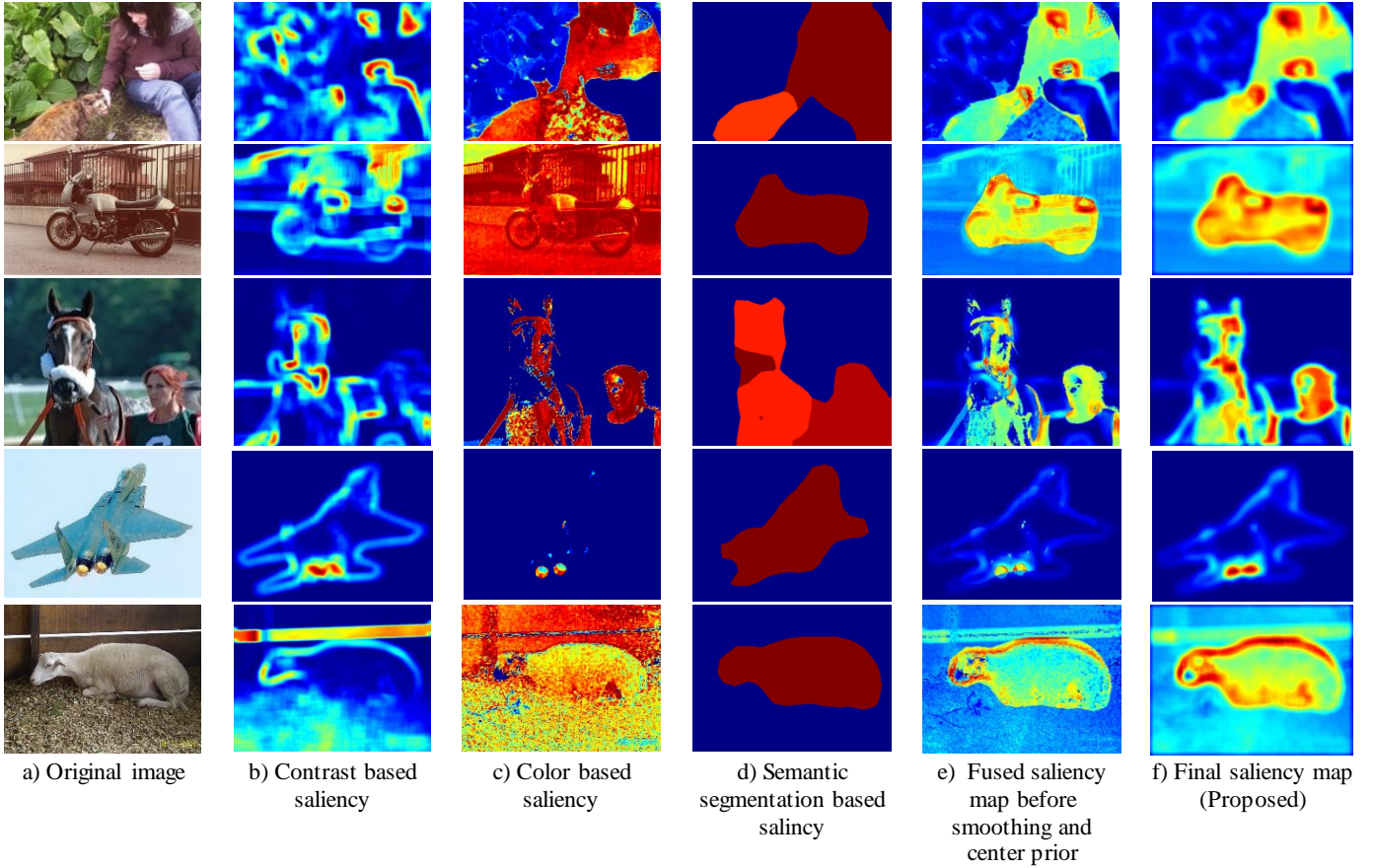

Figure. 6. Visual illustration of saliency map detection.

Table 2: Statistics of the segmentation image sets [12].

|  | train | | val | | trainval | | test | |
|---|---|---|---|---|---|---|---|---|
|  | img | obj | img | obj | img | obj | img | obj |
| Aeroplane | 88 | 108 | 90 | 110 | 178 | 218 | – | – |
| Bicycle | 65 | 94 | 79 | 103 | 144 | 197 | – | – |
| Bird | 105 | 137 | 103 | 140 | 208 | 277 | – | – |
| Boat | 78 | 124 | 72 | 108 | 150 | 232 | – | – |
| Bottle | 87 | 195 | 96 | 162 | 183 | 357 | – | – |
| Bus | 78 | 121 | 74 | 116 | 152 | 237 | – | – |
| Car | 128 | 209 | 127 | 249 | 255 | 458 | – | – |
| Cat | 131 | 154 | 119 | 132 | 250 | 286 | – | – |
| Chair | 148 | 303 | 123 | 245 | 271 | 548 | – | – |
| Cow | 64 | 152 | 71 | 132 | 135 | 284 | – | – |
| Diningtable | 82 | 86 | 75 | 82 | 157 | 168 | – | – |
| Dog | 121 | 149 | 128 | 150 | 249 | 299 | – | – |
| Horse | 68 | 100 | 79 | 104 | 147 | 204 | – | – |
| Motorbike | 81 | 101 | 76 | 103 | 157 | 204 | – | – |
| Person | 442 | 868 | 445 | 865 | 887 | 1733 | – | – |
| Pottedplant | 82 | 151 | 85 | 171 | 167 | 322 | – | – |
| Sheep | 63 | 155 | 57 | 153 | 120 | 308 | – | – |
| Sofa | 93 | 103 | 90 | 106 | 183 | 209 | – | – |
| Train | 83 | 96 | 84 | 93 | 167 | 189 | – | – |
| Tvmonitor | 84 | 101 | 74 | 98 | 158 | 199 | – | – |
| Total | 1464 | 3507 | 1449 | 3422 | 2913 | 6929 | – | – |

In Fig. 6 all steps of the proposed method are illustrated for some of the images of the Pascal-voc11 dataset. In Fig. 6(b) and Fig.6(c) salient points are resulted based on contrast and color information respectively. In Fig.6(d) salient points are based on semantic segmentation. In Fig. 6(e) results are from the fused saliency map and in Fig. 6(f) final results after performing center priors and mean smoothing filters are illustrated.

It can be observed that the fusion results have better representation of the salient points than the other three methods. In Fig. 7 the convergence of the MLP for context detection is depicted in terms of its accuracy.

Our results show that the accuracy of our method is highly related to semantic segmentation results. It is useful to note that we have used one of the earliest versions of deep neural networks for semantic segmentation. Using state-of-the-art CNN structures could lead to better semantic segmentations.

V. CONCLUSION

A novel algorithm in saliency detection based on semantic information was presented. Three saliency maps based on semantic features, color, and contrast were applied. A CNN was used to generate semantic segmentation labels and an MLP utilized these segmentations for context detection. Using High level image features in form of semantic segmentation yielded 99% accuracy in context detection. Simulation results, from the Pascal-voc11 image dataset, showed that using context detection as well as color and contrast features would lead to an acceptable saliency map.

In our future work we will use more accurate deep learning methods to produce semantic segmentation. Also we will consider more features to create complementary saliency maps.

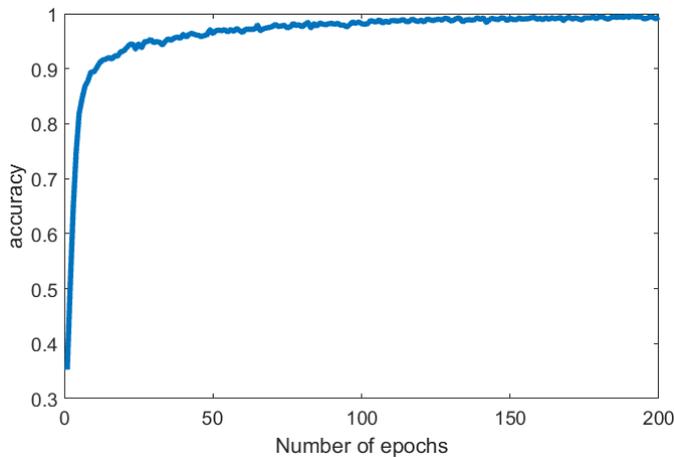

Figure. 7. Context detection training accuracy.

REFERENCES

[1] S. Ramanathan, H. Katti, N. Sebe, M. Kankanhalli and T.S. Chua, "An eye fixation database for saliency detection in images". *Computer Vision–ECCV,* pp.30-43, 2010,.

[2] H. Hadizadeh and I.V. Bajić, "Saliency-aware video compression," *IEEE Trans. Image Process.*, vol. 23, no. 1, pp. 19–33, 2014.

[3] J. Feng, Y. Wei, L. Tao, C. Zhang and J. Sun, "Salient object detection by composition," in *Proc. IEEE Int. Conf. Comput. Vis.*, Nov., pp. 1028–1035, 2011

[4] J. Park, J.-Y. Lee, Y.W. Tai and I.S. Kweon, "Modeling photo composition and its application to photo re-arrangement," in *Proc. IEEE Int. Conf. Image Process. (ICIP)*, pp. 2741–2744, 2012.

[5] W. Hu, R. Hu, N. Xie, H. Ling, and S. Maybank, "Image classification using multiscale information fusion based on saliency driven nonlinear diffusion filtering," *IEEE Trans. Image Process.*, vol. 23, no. 4, pp. 1513–1526, 2014.

[6] K. Wang, L. Lin, J. Lu, C. Li and K. Shi, "PISA: Pixelwise image saliency by aggregating complementary appearance contrast measures with edge-preserving coherence". *IEEE Transactions on Image Processing*, 24(10), 3019-3033, 2015.

[7] J. Kim, D. Han, Y. W. Tai and J. Kim, "Salient region detection via high-dimensional color transform and local spatial support". *IEEE transactions on image processing*, 25(1), 9-23, 2016.

[8] X. Xu, N. Mu, L. Chen, and X. Zhang, "Hierarchical salient object detection model using contrast-based saliency and color spatial distribution". *Multimedia Tools and Applications*, 75(5), 2667-2679, 2016.

[9] O. Muratov, P. Zontone, G. Boato and F.G. De Natale, (2011, May). "A segment-based image saliency detection". *IEEE International Conference on Acoustics, Speech and Signal Processing (ICASSP),* pp. 1217-1220, 2011.

[10] H. Li, J. Chen, H. Lu and Z. Chi, "CNN for saliency detection with low-level feature integration". *Neurocomputing*, 226, 212-220, 2017.

[11] J. Long, E. Shelhamer and T. Darrell, "Fully convolutional networks for semantic segmentation". *IEEE Conference on Computer Vision and Pattern Recognition* (pp. 3431-3440), 2015.

[12] The PASCAL Visual Object Classes Challenge (VOC2012), available at: http://pascallin.ecs.soton.ac.uk/challenges/VOC/voc2012/index.html.

[13] MatConvNet: CNNs for MATLAB, available at: http://www.vlfeat.org/matconvnet/pretrained/ .